\documentclass[10pt,twocolumn,letterpaper]{article}

\usepackage{wacv}
\usepackage{times}
\usepackage{graphicx}
\usepackage{amsmath}
\usepackage{amssymb}

\wacvfinalcopy 

\def\wacvPaperID{1015} 

\ifwacvfinal\fi
\begin{document}

\title{Eye Contact Correction using Deep Neural Networks}

\author{Leo F. Isikdogan, Timo Gerasimow, and Gilad Michael\\
Intel Corporation, Santa Clara, CA\\
{\tt\small \{leo.f.isikdogan, timo.gerasimow, gilad.michael\}@intel.com}}

\maketitle
\ifwacvfinal\thispagestyle{empty}\fi

\begin{abstract}
    In a typical video conferencing setup, it is hard to maintain eye contact during a call since it requires looking into the camera rather than the display. We propose an eye contact correction model that restores the eye contact regardless of the relative position of the camera and display. Unlike previous solutions, our model redirects the gaze from an arbitrary direction to the center without requiring a redirection angle or camera/display/user geometry as inputs. We use a deep convolutional neural network that inputs a monocular image and produces a vector field and a brightness map to correct the gaze. We train this model in a bi-directional way on a large set of synthetically generated photorealistic images with perfect labels. The learned model is a robust eye contact corrector which also predicts the input gaze implicitly at no additional cost.
    
    Our system is primarily designed to improve the quality of video conferencing experience. Therefore, we use a set of control mechanisms to prevent creepy results and to ensure a smooth and natural video conferencing experience. The entire eye contact correction system runs end-to-end in real-time on a commodity CPU and does not require any dedicated hardware, making our solution feasible for a variety of devices.
\end{abstract}

\section{Introduction}

Eye contact can have a strong impact on the quality and effectiveness of interpersonal communication. Previous evidence suggested that an increase in the amount of eye contact made by a speaker can significantly increase their perceived credibility~\cite{beebe1974eye}. However, a typical video conferencing setup creates a gaze disparity that breaks the eye contact, resulting in unnatural interactions. Many video-conferencing capable devices have a display and camera that are not aligned with each other. During video conferences, users tend to look at the other person on display or even a preview of themselves rather than looking into the camera. The gap between the camera and where the users typically look at makes it hard to maintain eye contact and have a natural, face to face, conversation.

Earlier solutions required specific hardware such as a pair of cameras that help synthesize gaze-corrected images~\cite{criminisi2003gaze,yang2002eye} or reflective screens similar to that of teleprompters. A more recent solution~\cite{kononenko2015learning} used a single camera to correct the gaze by 10-15 degrees upwards, assuming that a typical placement for a camera would be at the top-center of the device, just above the screen. However, many new portable devices have their cameras located at the top-left and top-right corners of the displays. Such devices would require horizontal gaze correction as well as the upwards correction. Furthermore, many tablets and smartphones can be rotated and used in any orientation. Different users may use their devices at different orientations and view the display from different distances. This effectively changes the relative position of the camera with respect to the user and the center of the display. Therefore, a universal eye contact corrector should support redirecting the gaze from an arbitrary direction to the center regardless of the relative camera and display positions.

\begin{figure}[t]
\centering
\includegraphics[width=0.38\linewidth]{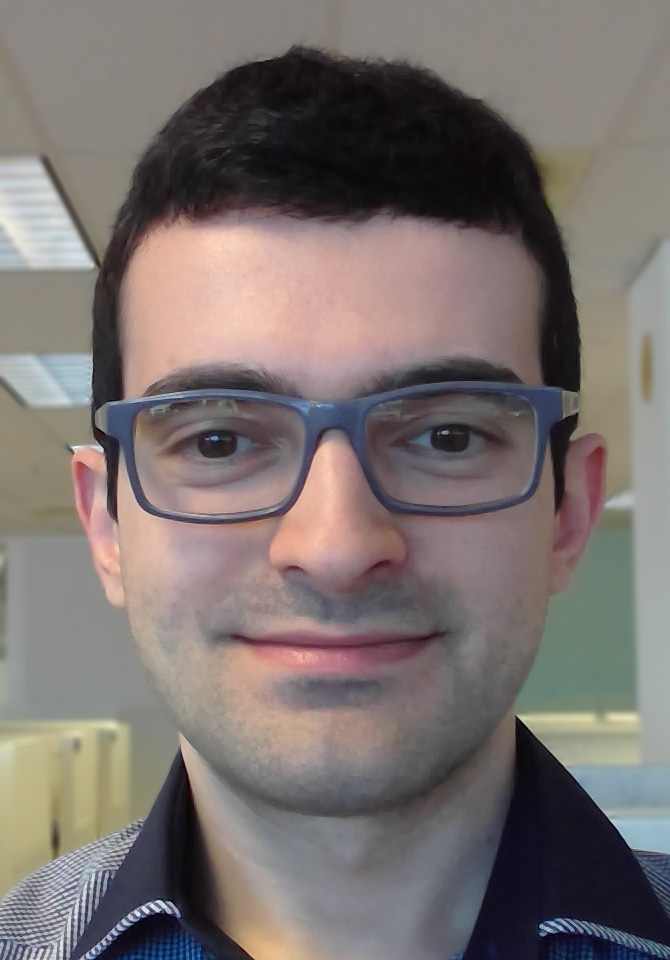}
\includegraphics[width=0.38\linewidth]{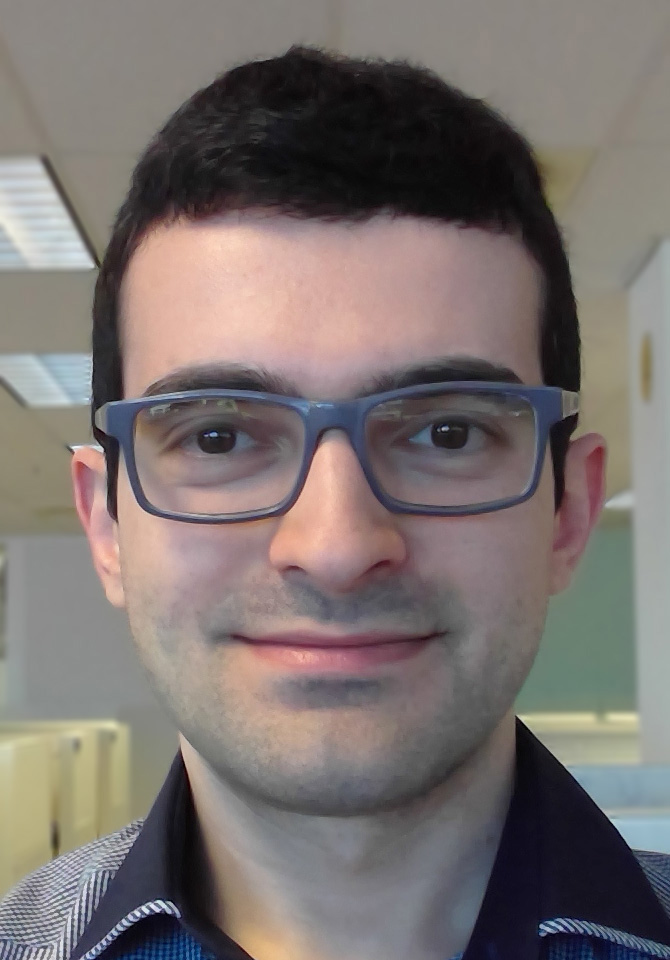}
\caption{Eye contact correction: the user is looking at the screen in the input frame (left). The gaze is corrected to look into the camera in the output frame (right).}
\label{fig:ecc_intro}
\end{figure}

A deep learning based approach~\cite{ganin2016deepwarp} showed that it is possible to redirect gaze towards an arbitrary direction, given a redirection angle. In a typical use case of eye contact correction, however, neither a redirection angle nor the input gaze direction is available. It is indeed possible to replace eyes with rendered 3D models of eyes to simulate an arbitrary gaze~\cite{wood2018gazedirector,shu2017eyeopener} without having a redirection angle. However, using such a model for gaze correction in video conferencing would be challenging since it is hard to render details such as eyelashes and glasses in real-time while remaining faithful to the original input.

We propose an eye contact correction system that is designed primarily to improve video conferencing experience. Our system first uses a facial landmark detector to locate and crop the eyes, and then feeds them into a deep neural network. Our proposed model architecture learns to redirect an arbitrary gaze to the center without requiring a redirection angle. We show that when a redirection angle is not given, the model learns to infer the input gaze implicitly. As a side product, our model predicts the input gaze direction and magnitude at no additional cost. Finally, our eye contact corrector outputs frames having smooth and naturally corrected gaze using a set of control mechanisms. Those mechanisms control the strength of the correction, prevent `creepiness' from overly corrected eye contact, and ensure temporal consistency in live applications. Our live application (Figure \ref{fig:ecc_intro}) runs in real-time on CPU, making our eye contact corrector a feasible solution for a wide range of devices.

\section{Related Work}
Eye contact correction can be considered a specific case of gaze manipulation where the gaze is redirected to the center in a video conferencing setup. Numerous solutions that specifically addressed the video conferencing gaze correction problem required additional hardware such as stereo cameras~\cite{criminisi2003gaze,yang2002eye} or depth sensors~\cite{kuster2012gaze,zhu2011eye,thies2018facevr}. Kononenko et al.~\cite{kononenko2015learning} proposed monocular solution that solely relied on images captured by a web camera. Their solution used ensembles of decision trees to produce flow fields, which are later used to warp the input images to redirect gaze upwards 10 to 15 degrees. As discussed earlier, this type of vertical correction works well only when the camera is located at the top center of the screen, with a predefined distance from the user. However, many hand-held devices can be used in both landscape and vertical orientations and at an arbitrary viewing distance.

A more flexible approach, named DeepWarp~\cite{ganin2016deepwarp}, used a deep neural network to redirect the gaze towards an arbitrary direction. DeepWarp can manipulate the gaze towards any direction, thus can be used for gaze correction in video conferencing regardless of device orientation and user distance, given a redirection angle as input. However, such a redirection angle is usually hard to obtain in real life scenarios. For example, even when the device type, orientation, and user distance is known, a fixed redirection angle would assume that all users look at the same point on the display to properly correct the gaze. In practice, windows that show the participants in a video call can be shown at different parts of the display. Furthermore, users may even prefer to look at the preview of themselves rather than the other person.

Wood et al.~\cite{wood2018gazedirector} proposed an approach that can redirect the gaze to any given direction without inputting a redirection angle. Their method created a 3D model of the eye region, recovering the shape and appearance of the eyes. Then, it redirected the gaze by warping the eyelids and rendering the eyeballs having a redirected gaze. However, the model fitting step in their algorithm limited the real-time capability of their approach.

Although some of the earlier work employed temporal smoothing techniques~\cite{kuster2012gaze}, earlier gaze correction and redirection solutions overall tried to correct the gaze constantly, without a control mechanism. Therefore, the use of a general-purpose gaze redirector for video conferencing would lead to unnatural results particularly when the user is not engaged or moves away from a typical use case.

\section{Data Preparation}
\begin{figure}[t]
\centering
\includegraphics[width=1.0\linewidth]{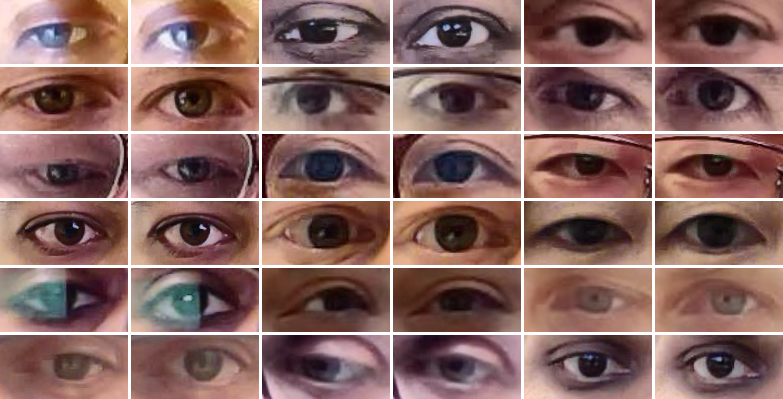}
\caption{Sample pairs from the natural dataset: the first image in every pair looks at a random point on the display whereas the second one looks into the camera.}
\label{fig:natural_data}
\end{figure}

\begin{figure}[t]
\centering
\includegraphics[width=1.0\linewidth]{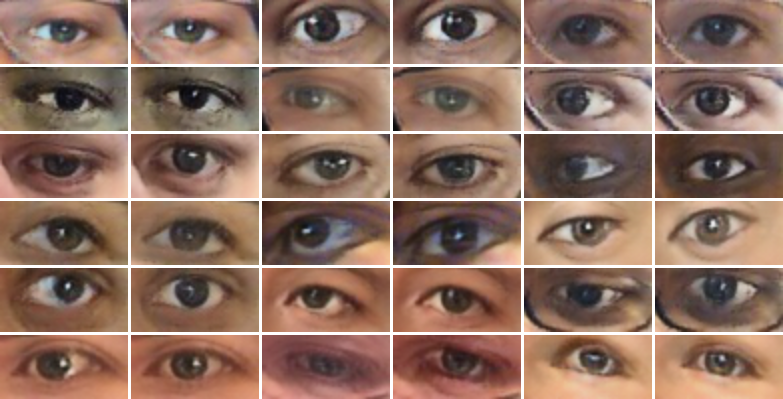}
\caption{Samples pairs from the synthetic dataset: each image pair belongs to a distinct randomized subject. The image pairs are aligned fixing everything except the gaze.}
\label{fig:synthetic data}
\end{figure}

To train and validate our system, we prepared two different datasets: one natural and one synthetic. The natural dataset (Figure \ref{fig:natural_data}) consists of image pairs where a subject looks into the camera and at a random point on display. Similarly, the synthetic dataset (Figure \ref{fig:synthetic data}) consists of image sets within which all factors of variation except for gaze stays constant. We used the natural dataset primarily to validate our model and to refine the samples in the synthetic dataset to look virtually indistinguishable from the natural ones. Being able to generate a photorealistic synthetic dataset allowed for generating an immense amount of perfectly-labeled data at a minimal cost.

\subsection{Natural Dataset}
We created a dataset that consists of image pairs where the participants saccaded between the camera and random points on display. The gaze of the participants was guided by displaying dots on the screen. The subjects participated in our data collection at their convenience without being invited into a controlled environment, using a laptop or tablet as the data collection device. Therefore, the collected data is representative of the typical use cases of the proposed application.

Unlike the gaze datasets that are collected in a controlled environment~\cite{smith2013gaze,fischer2018rt}, we did not use any apparatus to stabilize the participans' face and eyes or to prevent them from moving between frames. To locate the eyes in the captured frames, we used a proprietary facial landmark detector developed internally at Intel. The facial landmark detector provided a set of facial landmarks which we utilized to align and crop the eyes in the captured frames.

To improve the data quality, we created a routine that automatically deleted the frames that were likely to be erroneous. First, the cleaning routine removed the first frames in each sequence to compensate for any lagged response from the subjects. Second, it removed the frames where no faces were detected. Finally, it removed the frames where the subject was blinking, where the blinks were inferred from the distances between eye landmarks. We removed any incomplete pairs where either the input or ground truth images were missing to make sure all pairs in the dataset are complete. The clean dataset consisted of 3125 gaze pair sequences collected from over 200 participants.

\subsection{Synthetic Dataset}

Our synthetic data generator used the UnityEyes platform~\cite{wood2016learning} to render and rasterize images of eyes, which are later refined by a generative adversarial network. UnityEyes provides a user interface where the gaze can be moved by moving the cursor. We created sets of eye images by programmatically moving the cursor to move the gaze towards random directions. We modeled the cursor movements as a zero mean Gaussian random variable, where zero means a centered gaze, looking right into the camera.

To increase the diversity of samples in the dataset, we randomized subject traits, lighting, and head pose between different sets of images. We sampled 40 different gazes per set, where all images within a set had the same random configuration. Randomizing the subject traits changed the color, shape, and texture of the face, skin, and eyes. Using this process, we generated 3200 sets of artificial subjects with random traits, resulting in 128,000 images and nearly 2.5 million image pairs.

We limited the range of movement in the head pose randomization since we would not enable eye contact correction if the user is clearly looking at somewhere other than the camera and display. Therefore, we made sure that the head pose was within the limits of a typical use case where the eye contact correction would be practical to use. To further increase randomness, we also randomized the render quality of the synthesized images. Indeed, the use of highest possible render quality can be ideal for many applications. However, the amount of detail in those images, such as the reflection of the outside world on the surface of the eyes, can be unrealistic in some cases depending on the imaging conditions. After we captured raster images from the UnityEyes platform, we superimposed glasses of different sizes and shapes on some of the image sets. The glasses used 25 different designs as templates, where their size, color, and relative position were randomized within a visually realistic range.

UnityEyes provides facial landmarks for the eyes, which are comparable to the ones we used for the natural dataset.
Once the glasses are superimposed, we used those facial landmarks to align and crop the eyes. Since the images are generated synthetically, they can be perfectly aligned before the eyes are cropped. However, merely using a bounding box that encloses the eye landmarks leads to misaligned pairs. Cropping each image separately leads to small offsets between the images in the same set due to landmarks shifted by the gaze. Thus, we created a bounding box that fit all images in a given set and used a single bounding box per set. The bounding boxes had a fixed aspect ratio of 2:1 and are padded to have twice as much width as the average width in a given set.

To enhance photorealism, we used a generative adversarial network that learned a mapping between synthetic and real samples and brought the distribution of the synthetic data closer to real ones. Using the trained generator, we refined all images in the synthetic dataset to create a large dataset that consists of photorealistic images having virtually perfect labels. This process is detailed in Section \ref{sec:experiments}.

\begin{figure*}[t]
\centering
\includegraphics[width=1.0\linewidth]{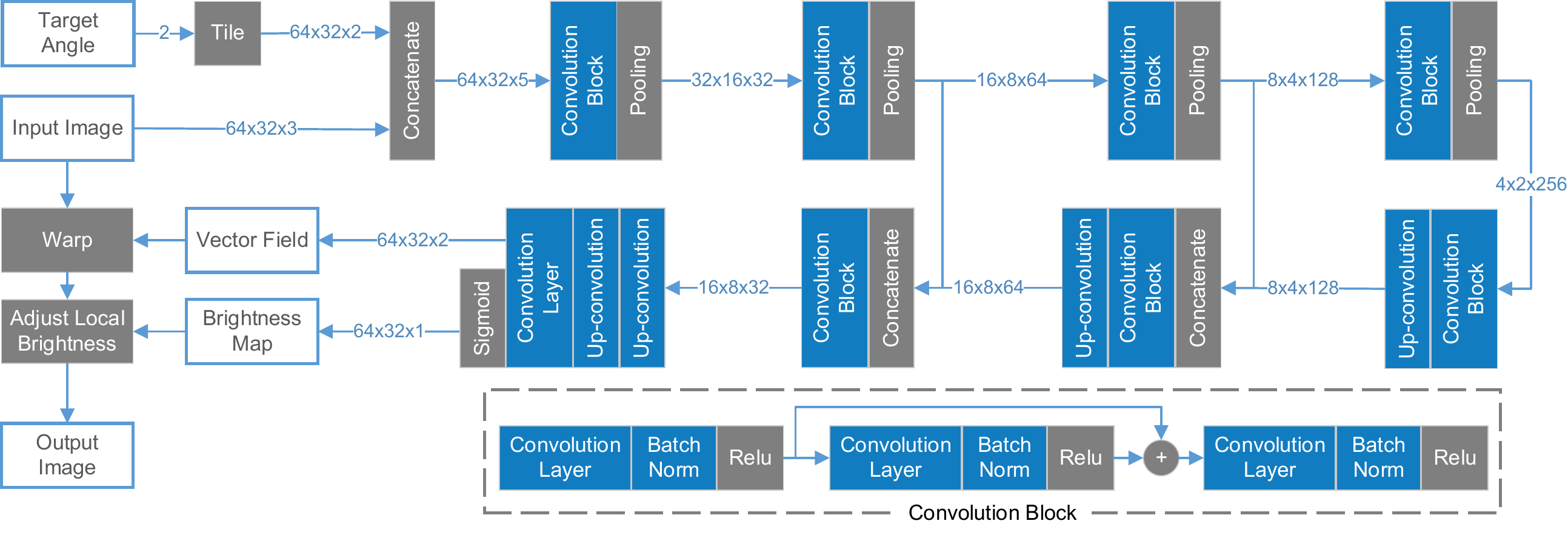}
\caption{The architecture of the eye contact correction model: ECC-Net inputs a patch that contains a single eye, warps the input to redirect gaze, and adjusts the local brightness to enhance eye clarity. Blocks with trainable parameters are shown in blue.}
\label{fig:eccnet_architecture}
\end{figure*}

All of the steps mentioned above are done only once as a pre-processing step. The pre-processed image pairs are also distorted on the fly during training with additive noise, brightness and contrast shift, and Gaussian blur, in random order and magnitude. These distortions not only emulate imperfect imaging conditions but also further augment the diversity of the samples in the dataset.

\section{The ECC-Net Model}
Our eye contact correction model, named ECC-Net, inputs an image patch that contains a single eye and a target gaze vector. The image patches are resized to $64 \times 32$ before they are fed into the model. The target gaze vector is represented in the Cartesian domain with its horizontal and vertical components and is tiled to have the same spatial dimensions as the input image. Once the training is complete, the target angle is set to zeros to redirect the gaze to center.

The core of ECC-Net is a fully-convolutional encoder-decoder network which uses U-Net style skip connections and channel-wise concatenations~\cite{ronneberger2015u} to recover details lost at the pooling layers. The model does the bulk of processing in low resolution both to reduce the computational cost and to improve spatial coherence of the results. The convolutional blocks in the model consist of three depthwise-separable convolutional layers with a residual connection~\cite{resnet} that skips over the middle layer. The convolutional layers use batch normalization \cite{ioffe2015batch} and ReLU activations.

The model produces a flow field and a brightness map similar to the methods presented in~\cite{kononenko2015learning} and~\cite{ganin2016deepwarp}. The output layer consists of two up-convolution layers ($2 \times 2$ convolution with a stride of $1/2$) followed by a convolutional layer having a 3-channel output. Two of these channels are used directly to predict the horizontal and vertical components of a vector field that is used to warp the input image. The third channel is passed through a sigmoid function and used as a map to adjust local brightness. Using such a mask is shown to be effective in improving the appearance of eye whites after gaze warping~\cite{ganin2016deepwarp}. The brightness mask enhances eye clarity and corrects the artifacts that result from horizontal warping when there are not enough white pixels to recover the eye white. Both the warping and brightness adjustment operations are differentiable. Therefore, the entire network can be trained end-to-end. The overall architecture of the model is shown in Figure \ref{fig:eccnet_architecture}.

For eye contact correction, training a model to output a vector field has several advantages over training a generic encoder-decoder model that produces pixel-wise dense predictions. First, the vector fields produced by the model can be easily modified in a meaningful way using external signals. For example, their magnitude can be scaled before warping to control the correction strength. Those vectors can also be averaged over time for temporal smoothing without producing blurry results (Section \ref{sec:control}). Second, predicting a motion vector imposes the prior that pixels should move rather than changing in an unconstrained way when the gaze changes. Finally, training a model to output the pixel values directly can lead to a bias towards the mean image in the training set~\cite{ganin2016deepwarp}, resulting in loss of detail.

Indeed, images can be generated with a high level of detail using an adversarial loss~\cite{lotter2015unsupervised} instead of a mean squared error loss. A generative adversarial network (GAN) can learn what is important to produce in the output~\cite{goodfellow2014generative}. However, although generative adversarial networks are better at reconstructing details, the details they produce might originate neither in the input nor the ground truth. A model that is trained with an adversarial loss can hallucinate details when the output is comprised of unrestricted pixels. This behavior might be acceptable or even preferred for many applications. However, we would not want this type of flexibility to redirect gaze in a video conferencing setup. For example, adding eyelashes or any other traits that are hallucinated might lead to unnatural results. Therefore, we built a model that manipulates the location and brightness of existing pixels. This approach ensures that any detail that is in the output originates in the input. The applicability of this approach would be limited in some extreme cases, such as when the input image is a nearly closed eye. In those cases, we smoothly disable the model to avoid any unnatural correction.

\section{Bi-directional Training}
\begin{figure}[t]
\centering
\includegraphics[width=1.0\linewidth]{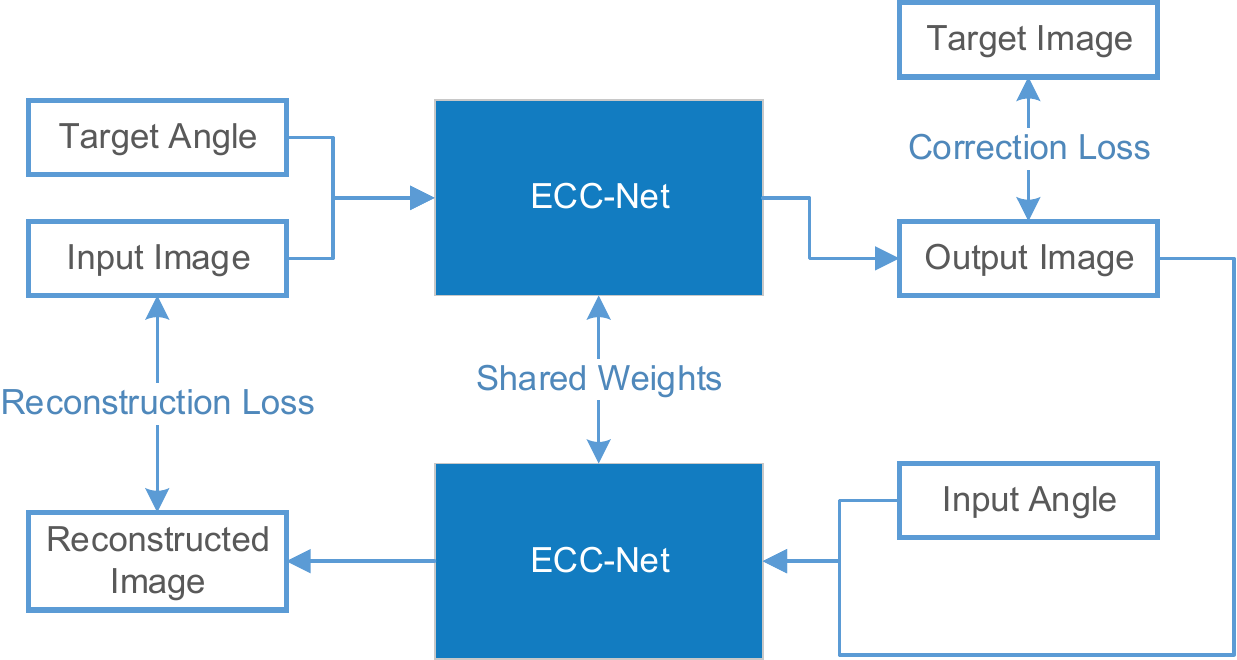}
\caption{Bi-directional training: the model optimizes the correction and reconstruction losses concurrently to enforce mapping reversibility.}
\label{fig:cyclic_training}
\end{figure}

We trained ECC-Net in a bi-directional fashion to enforce mapping reversibility. The model is first given an input image and a target angle to redirect the gaze. In the first direction, the model is expected to minimize a correction loss $L_{c}$, which is defined as the mean squared error between the gaze-corrected and ground truth images. In the other direction, the model is given the gaze-corrected output image and the input angle to redirect the gaze back to its original state. Although this should be the expected behavior of a gaze redirection model, we found that some warping artifacts in the output make it difficult to recover the original image. To address this problem, we defined a reconstruction loss $L_{r}$ between the reconstructed image and the original image and optimize it concurrently with the correction loss (Figure \ref{fig:cyclic_training}). This type of cycle-consistent loss has been previously used to learn unpaired image-to-image translations in generative adversarial networks~\cite{zhu2017unpaired}.

Training the model in a bi-directional way reduced the artifacts and resulted in more natural gaze redirection results. However, assigning the correction and reconstruction losses the same weight during training led to a mode collapse where the model quickly converged to an identity transform to minimize the reconstruction loss. Readjusting the weights of the losses in the total loss function as $L_{total} = 0.8 L_{c} + 0.2 L_{r}$ helped the optimizer keep a good balance between the loss functions in both directions.

The target angles are used only during training and set to $(0,0)$ during inference since the goal of the model is to move the gaze to the center. Using target angles other than zero during training improved the robustness of the model and allowed for post-training calibration. For example, if the gaze is still off after correction on a particular device then the target angle can be tuned to compensate for the offset, although this should not be necessary in a typical case. Using pairs of images having arbitrary gazes also increased the number of possible image pairs in the training data. For example, using a set of 40 images for a given subject, ${40 \choose 2} = 780$ unique pairs can be generated as compared to $39$ pairs using a single target. This effectively augmented the data and reduced the risk of overfitting.

\section{Gaze Prediction}
\begin{figure}[t]
\centering
\includegraphics[width=0.85\linewidth]{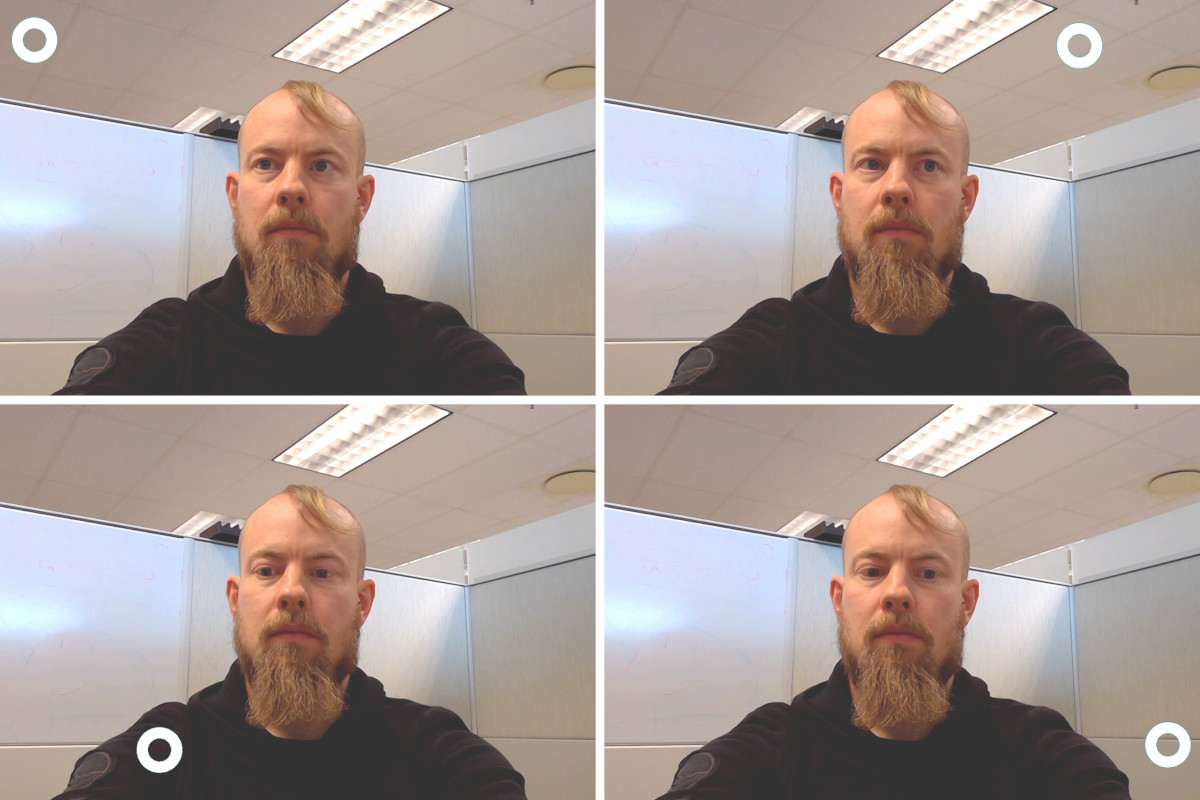}
\caption{Gaze prediction: ECC-Net predicts the input gaze as a byproduct of eye contact correction. The white circle in the figure shows the predicted gaze. See the demo video for more results.}
\label{fig:gaze_prediction}
\end{figure}

An intriguing phenomenon we observed is that the model learned to predict the input gaze implicitly. We found that computing the mean motion vector, negating its direction, and scaling its magnitude to fit the screen gives a coarse estimate of the input gaze (Figure \ref{fig:gaze_prediction}). Unlike a typical multi-task learning setup where a model is trained to perform multiple tasks simultaneously, our model learns to perform two tasks while being trained to perform only one of them. Therefore, we can arguably consider the eye contact correction problem as a super-set of gaze prediction, at least in a loose sense. A model should have an internal representation of the input gaze to be able to manipulate it. However, this representation need not be as comprehensive as a dedicated gaze prediction model would learn.

We should note that our model is not a fully-blown gaze predictor, but rather is an eye contact corrector that learns the input gaze to function better. This behavior is likely a byproduct of training the model to redirect gaze without providing a redirection angle, which requires the input gaze angle to be inferred. The inferred gaze does not incorporate head pose or distance from the screen and relies only on the information extracted from eyes in isolation. Therefore, it should not be expected to be as accurate as systems that use dedicated sensors or models~\cite{cvpr2016_gazecapture, sugano2017s, fischer2018rt, Wang_2018_CVPR} that are designed specifically for gaze prediction.

We make no claims on the gaze prediction accuracy of our model as we do not benchmark it against state-of-the-art gaze predictors. However, the gaze predicted by our model as a side product can still be practical to use in a variety of use cases, especially where the computational cost is a concern. The additional cost of computing a coarse estimate of the input gaze, i.e., mean computation and negation, is virtually negligible. Therefore, a video conferencing application that already uses our eye contact corrector would be able to compute gaze statistics with minimal overhead. The real-time gaze information would also enable hands-free interactions, such as dimming the backlight when the user is not engaged. Thus, the gaze prediction property of our eye contact corrector has the potential to decrease battery consumption while providing additional functionality.

\section{Control Mechanism}
\label{sec:control}
We provide a set of mechanisms that control the correction strength smoothly to ensure a natural video conferencing experience. The control mechanisms we use can be grouped into two blocks: a control block that reduces the correction strength by scaling the ECC-Net output when needed, and a temporal stability block that temporally filters the outputs.

\begin{figure}[t]
\centering
\includegraphics[width=0.8\linewidth]{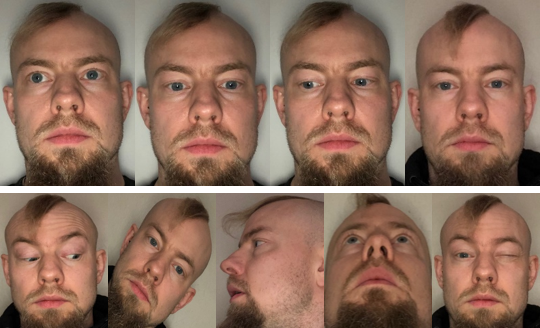}
\caption{Control mechanism: ECC is enabled for typical use cases (top) and disabled when the user diverts away from the primary use case (bottom).}
\label{fig:ecc_control}
\end{figure}

Eye contact correction is disabled smoothly when the user is too far from the center, too far away from the screen, too close to the screen, or blinking. The correction is also disabled when the user looks somewhere other than the camera and display (Figure \ref{fig:ecc_control}). The control block monitors the face size, distance from the center, head pose (i.e., pitch, roll, yaw), and eye opening ratio, which are inferred from the output of the same facial landmark detector that we use to align and crop the eyes. In addition to the facial landmarks, the control block also factors in mean and maximum motion vector magnitudes to limit correction for extreme gazes. Both landmark and motion vector based signals produce a scaling factor between 0 and 1. The overall correction strength is calculated by multiplying those scaling factors calculated for each triggering signal.

We tuned the parameters for each of those signals by performing subjective quality tests on a group of 100 people. We chose a configuration where most users did not realize when the control mechanisms enable or disable eye contact correction but felt like they had eye contact. We also quantified the value of the control mechanisms by measuring their impact on the test set error of the system (Section~\ref{sec:experiments}).

The stability block filters the motion vectors temporally using an alpha-beta filter, which is a derivative of the Kalman filter~\cite{penoyer1993alpha}. The filtering is done on the vector field before warping input images rather than pixel values after warping. This process eliminates flicker and outlier motion vectors in an input video stream without blurring out the output images. When used together with the control block, the temporal stability block ensures the eye contact correction operates smoothly in a video conferencing setting.

Overall, the control mechanisms prevent abrupt changes and ensure that the eye contact corrector avoids doing any correction when the user diverts away from a typical video conferencing use case. Consequently, the eye contact corrector operates smoothly and prevents `creepy' or unneeded corrections.

\section{Experiments}
\label{sec:experiments}
In our experiments, we trained ECC-Net using only the synthetic dataset and used the natural dataset as a validation set to pick the best performing model configuration. Once the training is complete, we tested the frozen model on the Columbia Gaze Dataset~\cite{smith2013gaze}, which is a public benchmark dataset that was originally used for eye contact detection~\cite{smith2013gaze}. We reorganized the Columbia Gaze Dataset to have gaze pairs similar to our natural dataset. Using data from entirely different sources for training, validation, and test sets minimized the risk of overfitting, including its implicit forms such as information leakage from the validation set due to excessive hyperparameter tuning or dataset bias~\cite{torralba2011unbiased}.

Initially, we trained the model on both left and right eyes, where left eyes on the synthetic dataset were generated by flipping right eyes. This resulted in a poor horizontal correction since the model needed to put considerable effort to decide whether the input is a left or right eye to be able to correct the gaze horizontally in the right amount. To better utilize the model capacity for correction, we trained the model to operate on the right eyes only by flipping the left eyes during inference. Consequently, the model learned to correct the gaze better both horizontally and vertically.

We used the relative reduction in mean squared error as the performance metric and modified it to be more robust against minor misalignments. This misalignment-tolerant error used the minimum of errors between image pairs shifted within a slack of 3x3 pixels. We found the misalignment-tolerant error more consistent with the visual quality of the results as compared to a rigid pixel-to-pixel squared error.

We trained our model for about 3 million iterations, using an Adam \cite{kingma2014adam} solver with $\beta_1=0.9$ $\beta_2=0.999$, $\epsilon=0.1$, and a cyclic learning rate \cite{smith2017cyclical} between 0.002 and 0.01. Using a relatively large $\epsilon$ helped stabilize training. The error reached its minimum value at around 2 million iterations. We were able to further decrease the overall error by using a portion of the natural dataset for fine-tuning and the rest for validation. We tuned only the first layers (layers before the first skip connection) while the rest of the network stayed frozen, which can be considered a form of domain adaptation. Using a portion of the natural data for fine tuning decreased the error marginally.

\begin{figure}[t]
\centering
\includegraphics[width=0.84\linewidth]{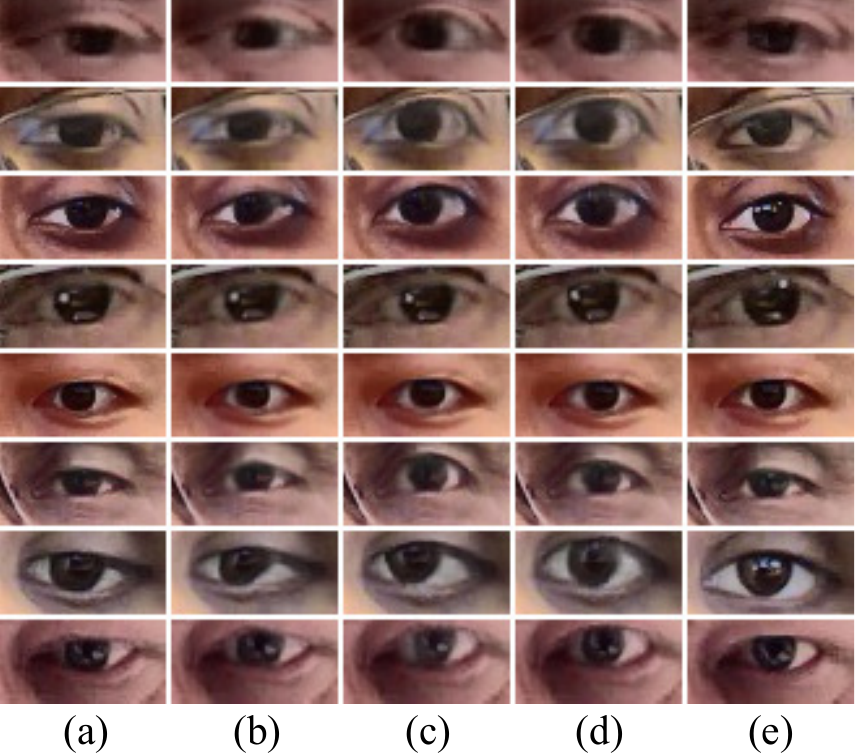}~
\caption{Results on samples from the validation set: (a)~input, (b)~model fine-tuned on natural data, (c)~model trained on unrefined synthetic data only, (d)~model trained on refined synthetic data, (e)~ground truth.}
\label{fig:test_results_natural}
\end{figure}

Although fine tuning on natural data helped reduce the error, it also noticeably decreased the correction strength and worsened the qualitative results (Figure \ref{fig:test_results_natural}). Despite the misalignment-tolerant error metric, some of the remaining error on the natural dataset was still due to the differences other than the gaze, such as shadows and reflections. We observed that a substantial decrease in the error was a result of better gaze correction whereas smaller `improvements' were a result of closer-to-average results that smoothed out other factors of variation. Therefore, we used the natural dataset as a development set and calculated the error as a sanity check rather than as a benchmark, while continuously monitoring the results qualitatively. Overall, training the model solely on synthetic data resulted in visually better results. This is likely a result of the impact of perfect labels in the synthetic set outweighing the impact of a data distribution closer to the real use case in the natural set.

\begin{figure}[t]
\centering
\includegraphics[width=1.0\linewidth]{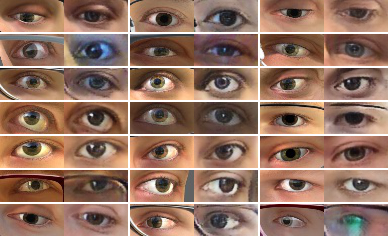}
\caption{Samples from the synthetic dataset before (left) and after (right) they are refined using a generator network. The refined images reveal some details about the distribution of data in the natural dataset, such as reflections in the eyes and glare on glasses. The generator brings the distribution of the synthetic data closer to real data and makes eyes and their surroundings more photorealistic by adding those details among many others.}
\label{fig:natural_data_refined}
\end{figure}

\begin{table}[t]
\begin{center}
\begin{tabular}{l||cc}
\hline
\textbf{Model and Training Data} & \textbf{Val. Error} & \textbf{Test Error} \\ \hline
ECC-Net \\ Unrefined Synthetic &    0.386    &     0.431  \\ \hline
ECC-Net \\ Natural + Synthetic &    \textbf{0.372}    &     0.465  \\ \hline
ECC-Net \\ Refined Synthetic   &    0.375    &     \textbf{0.414}  \\ \hline
DeepWarp \cite{ganin2016deepwarp} (Baseline) \\ Refined Synthetic &    0.411    &     0.442  \\ \hline
\end{tabular}
\end{center}
\caption{The relative mean squared error on the validation (natural dataset) and test (Columbia Gaze) sets when the model is trained on synthetic data before and after refinement. Training a model on refined synthetic images achieved a similar error as training it on unrefined images followed by fine tuning on a disjoint portion of the natural dataset. However, the model that used the refined synthetic data achieved a low error via better gaze correction whereas the model that is fined tuned on the natural data produced closer to average results (Figure \ref{fig:test_results_natural}). ECC-Net achieved a lower error than the baseline model, given the same training setup.}
\label{table:test_results}
\end{table}

\begin{figure*}[t]
\centering
\includegraphics[width=1.0\linewidth]{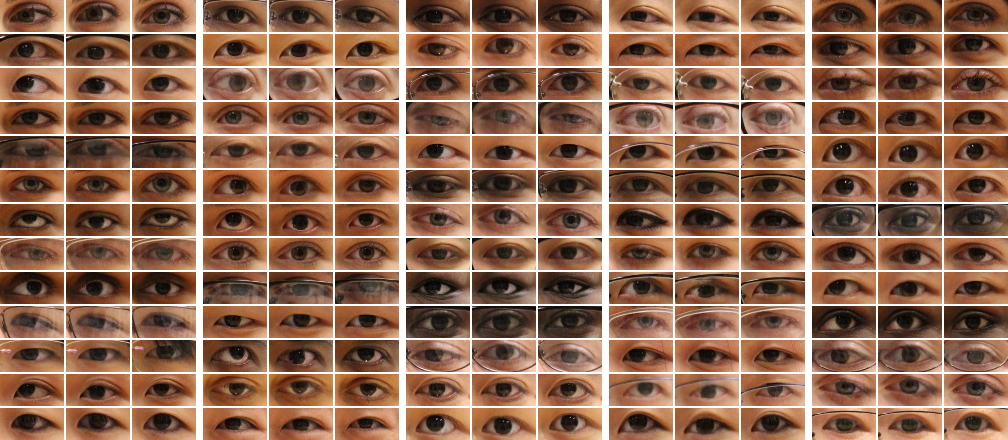}
\caption{Results on a random subset of the Columbia Gaze Dataset. The model is trained using only synthetic samples which were refined using the natural images in our dataset. The leftmost image in each group shows the input, the middle image shows the ECC-Net result, and the rightmost image shows the ground truth which was used to compute the test set error.}
\label{fig:test_results_columbia}
\end{figure*}

To bring the distribution of the synthetic data closer to real data without sacrificing the label quality, we built a generative adversarial network based on CycleGAN~\cite{zhu2017unpaired}. CycleGAN uses a cycle-consistent training setup to learn a mapping between two image sets without having a one-to-one correspondence. We modified and trained CycleGAN to learn a mapping between our synthetic and natural datasets, generating a photorealistic eye image given a synthetic sample. In our training setup, we used two additional mean absolute error (L1) losses defined between the inputs and outputs of the generators to further encourage input-output similarity. This type of `self-regularization' loss has been previously shown to be effective for training GANs to refine synthetic images~\cite{shrivastava2017simgan}. We defined the additional loss functions only on the luminance channel to give the model more flexibility to modify color while preserving the gaze direction and the overall structure of the input. We used the default hyperparameters for CycleGAN for training, treating the additional L1 losses the same as the reconstruction losses. The trained generator produced photorealistic images without changing the gaze in the input (Figure \ref{fig:natural_data_refined}).

Training the model on the GAN-refined synthetic images achieved a similar error as the fine tuned model without degrading the visual quality of the outputs. The results had almost no artifacts for the typical use cases. The artifacts were minimal even for the challenging cases such as where there is glare, glass frames are too close to the eye, or the scene is too dark or blurry. Qualitative results on a random subset of the Columbia Gaze Dataset are shown in Figure \ref{fig:test_results_columbia}. The visual examples show that some of the error between the gaze-corrected and ground truth images is explained by factors other than the gaze, such as moved glass frames and hair occlusions. The results look plausible even when they are not very similar to the ground truth images.

As a baseline, we trained the DeepWarp~\cite{kononenko2015learning} model using the same training setup. Both ECC-Net and DeepWarp models resulted in a mean squared error that is significantly lower than identity transform (59\% and 56\% reduction in test error, respectively). ECC-Net performed better than DeepWarp despite not requiring redirection angles as inputs. The results are summarized in Table \ref{table:test_results}.

We disabled the control mechanisms in the experiments detailed above. However, in the validation and test sets, we excluded the images where the control block would disable the system entirely. To quantify the impact of the control mechanisms, we also performed an ablation study where we included those images in the validation and test sets. Filtering out the images that triggered the control mechanisms significantly reduced the error on both sets (Table \ref{table:control_mechanisms}).

\begin{table}[t]
\begin{center}
\begin{tabular}{l||cc}
\hline
\textbf{Control Mechanisms} & \textbf{Val. Error} & \textbf{Test Error} \\ \hline
Enabled  &  0.375    &     0.414  \\ \hline
Disabled &  0.423    &     0.442  \\ \hline
\end{tabular}
\end{center}
\caption{The impact of the control mechanisms on the error on validation and test sets.}
\label{table:control_mechanisms}
\end{table}

\section{Conclusion}
We presented an eye contact correction system that redirects gaze from an arbitrary angle to the center. Our eye contact corrector consists of a deep convolutional neural network, which we call ECC-Net, followed by a set of control mechanisms. Unlike previous work, ECC-Net does not require a redirection angle as input, while inferring the input gaze as a byproduct. It supports a variety of video conferencing capable devices without making an assumption about the display size, user distance, and camera location. ECC-net preserves details, such as glasses and eyelashes, without hallucinating details that do not exist in the input. Our training setup prevents destructive artifacts by enforcing mapping reversibility. The trained model employs control mechanisms that actively control the gaze correction during inference to ensure a natural video conferencing experience. Our system improves the quality of video conferencing experience while opening up new possibilities for a variety of other applications. Those applications may include software teleprompters and personal broadcasting applications that provide cues on display while maintaining eye contact with the viewers.

{\small
\bibliographystyle{ieee}

\vfill
\section*{Disclaimer}
No license (express or implied, by estoppel or otherwise) to any intellectual property rights is granted by this document. This document contains information on products, services and/or processes in development. All information provided here is
subject to change without notice. Intel and the Intel logo are trademarks of Intel Corporation in the U.S. and/or other
countries. \\ {\small\textcopyright}  Intel Corporation.

}

\end{document}